%% file: main.tex
\documentclass[conference]{IEEEtran}

\newcommand*{\ARXIV}{}

\newcommand*{\TIKZDIAGRAM}{}

\ifdefined\ARXIV
    \IEEEoverridecommandlockouts
    \usepackage{multicol}
    
    \newcommand\blfootnote[1]{%
      \begingroup
      \renewcommand\thefootnote{}\footnote{#1}%
      \addtocounter{footnote}{-1}%
      \endgroup
    }
    \IEEEaftertitletext{\vspace{-2.3\baselineskip}}
\fi

\usepackage{amsmath}    
\usepackage{amssymb}    
\usepackage{cite}       
\usepackage{textcomp, gensymb}    
\usepackage{graphicx}   
\usepackage[pagebackref,breaklinks,colorlinks]{hyperref}  
\usepackage{mathtools}  
\usepackage{stmaryrd}   
\usepackage{svg}        

\usepackage{graphicx}
\usepackage{subfig}
\usepackage{lipsum}
\usepackage{afterpage}

\makeatletter
\newcommand{\setcaptype}[1]{\def\@captype{#1}}
\makeatother
\newsavebox{\tempbox}

\begin{document}
\title{Formal and Practical Elements \\for the Certification of Machine Learning Systems\ifdefined\ARXIV\vspace*{-0.2\baselineskip}\fi}

\author{
    \IEEEauthorblockN{Jean-Guillaume Durand\IEEEauthorrefmark{1} and Arthur Dubois\IEEEauthorrefmark{2}}
    \IEEEauthorblockA{
        \IEEEauthorrefmark{1}Head of Perception, \texttt{jgd@xwing.com}\\
        \IEEEauthorrefmark{2}Director of Engineering, \texttt{arthur.dubois@xwing.com}\\
        Xwing, San Francisco, CA 94102
    }
\and
    \IEEEauthorblockN{Robert J. Moss}
    \IEEEauthorblockA{
        PhD Student, Department of Computer Science\\
        \texttt{mossr@cs.stanford.edu}\\
        Stanford University, Stanford, CA 94305}
}

\maketitle

\ifdefined\ARXIV
    \begin{multicols}{1}
        \blfootnote{\copyright 2023 IEEE.  Personal use of this material is permitted.  Permission from IEEE must be obtained for all other uses, in any current or future media, including reprinting/republishing this material for advertising or promotional purposes, creating new collective works, for resale or redistribution to servers or lists, or reuse of any copyrighted component of this work in other works.}
    \end{multicols}
\fi

\savebox{\tempbox}{
    \begin{minipage}{\textwidth}
        \setcaptype{figure}
        \centering
        \footnotesize  
        \ifdefined\TIKZDIAGRAM
            \includegraphics[width=\linewidth]{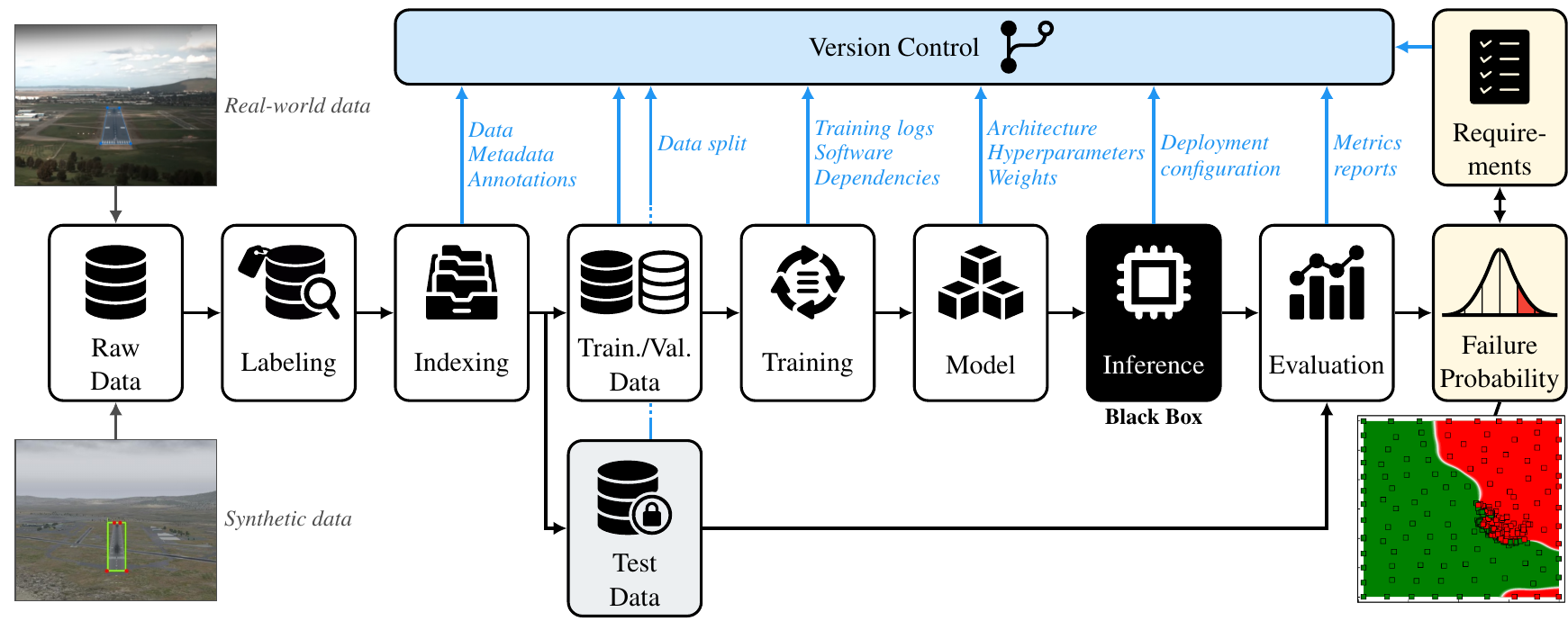}
        \else
            \includesvg[inkscapelatex=false,width=\linewidth]{img/steps.svg}
        \fi
        \caption{Overview of the proposed framework.}
        \label{fig:steps}
        \ifdefined\ARXIV
            \ifdefined\TIKZDIAGRAM
                \vspace*{-0.3\baselineskip}
            \else
               \vspace*{-2.3\baselineskip}
            \fi
        \else
            \ifdefined\TIKZDIAGRAM
                \vspace*{0.5\baselineskip}            
            \fi
        \fi
    \end{minipage}
}

\begin{figure}[t]
    \rlap{\usebox\tempbox}
\end{figure}
\afterpage{
    \begin{figure}[t]
    \rule{0pt}{\dimexpr \ht\tempbox+\dp\tempbox}
    \end{figure}
}

\ifdefined\ARXIV\vspace*{-2\baselineskip}\fi
\begin{abstract}
Over the past decade, machine learning has demonstrated impressive results, often surpassing human capabilities in sensing tasks relevant to autonomous flight. Unlike traditional aerospace software, the parameters of machine learning models are not hand-coded nor derived from physics but learned from data. They are automatically adjusted during a training phase, and their values do not usually correspond to physical requirements. As a result, requirements cannot be directly traced to lines of code, hindering the current bottom-up aerospace certification paradigm.
This paper attempts to address this gap by 
1) demystifying the inner workings and processes to build machine learning models,
2) formally establishing theoretical guarantees given by those processes,
and 3) complementing these formal elements with practical considerations to develop a complete certification argument for safety-critical machine learning systems. Based on a scalable statistical verifier, our proposed framework is model-agnostic and tool-independent, making it adaptable to many use cases in the industry. We demonstrate results on a widespread application in autonomous flight: vision-based landing.
\end{abstract}

\section{Introduction}
\label{sec:introduction}
\input{sections/00_introduction.tex}

\section{Related Work}
\label{sec:related_work}
\input{sections/01_related_work.tex}

\section{Proposed approach}
\label{sec:proposed_approach}
\input{sections/02_proposed_approach.tex}

\section{Results}
\label{sec:results}
\input{sections/03_results.tex}

\section{Conclusion}
\label{sec:conclusion}
\input{sections/04_conclusion.tex}

\newpage
\bibliographystyle{./bibtex/IEEEtran}
\bibliography{./bibtex/IEEEabrv,references.bib}

\end{document}

%% file: sections/00_introduction.tex
Machine learning (ML) is a long-established field that builds \textit{input-output} models by automatically learning from example data. Over the past decade, a succession of seminal research breakthroughs~\cite{dean2022golden} rekindled the field and unlocked a wide range of new applications.
\ifdefined\ARXIV\blfootnote{\vspace*{13.95pt}}\fi
That progress was largely possible due to increased amounts of available data and computational power. Deep learning is a subset of machine learning, with a subtle difference. In machine learning, the model learns from data features which are handcrafted by subject matter experts (i.e., object color, edges, texture, velocity, etc.). In deep learning, the model learns these features on its own. For many specialized tasks, such models have outperformed traditional methods, and even human specialists. These tasks range from medicine diagnosis to satellite imagery analysis, disease classification, board and video games strategies, and many others. Researchers and practitioners regard deep learning as a key enabler for autonomy in robotics, allowing unprecedented levels of safety. In particular, deep learning has been behind the autonomous driving successes of the past decade, and has the potential to unlock the true benefits of autonomous flight~\cite{durand2022potential}. In spite of these achievements, the widespread adoption of machine learning has been slow for safety-critical systems such as cars and aircraft. This is largely because of a lag in regulations and the absence of standardized certification policies, especially in aerospace.

\begin{figure}[!ht]
    \centering
    \includesvg[inkscapelatex=false,width=\linewidth]{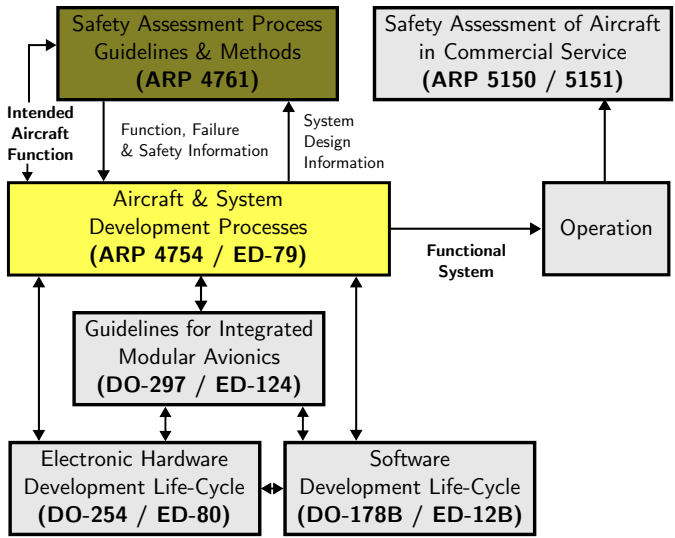}
    \caption{ARP4754A framework~\cite{arp4754a2010guidelines}}
    \label{fig:arp_4754a}
    \vspace*{2mm}
\end{figure}

In aerospace, certification aims at ensuring that hardware, software, and processes follow a defined set of design and performance requirements, and ultimately have a very low probability of failure. The key to obtaining operational approval lies in quantitatively demonstrating reliability over an entire Operational Domain (OD). For aerospace, government agencies grant certification, such as the Federal Aviation Administration (FAA) in the US, and the European Union Aviation Safety Agency (EASA) in Europe. They do not set exact practices to follow but rather issue advisory circulars to recognize acceptable means for developing and certifying aircraft, aerospace systems, or components.

These agencies are supported by professional associations such as SAE International, RTCA Inc., or the European Organisation for Civil Aviation Equipment (EUROCAE). They provide a forum to devise technical standards and recommended practices for the design of aircraft systems. These documents do not carry any legal force but are often referenced as acceptable means of compliance. The regulatory organization then works with each applicant to approve the use of each standard to define the formal Means of Compliance (MOC).

\begin{figure}[!ht]
    \centering
    \includesvg[inkscapelatex=false,width=\linewidth]{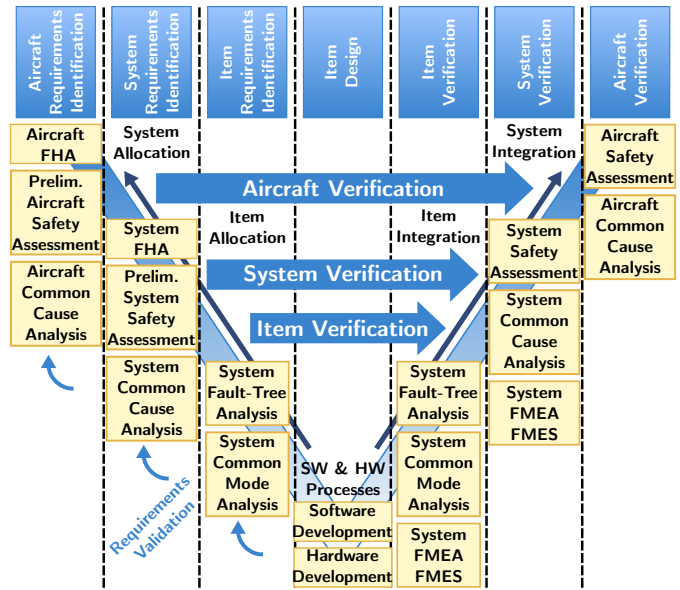}
    \caption{ARP4754A V-lifecycle~\cite{arp4754a2010guidelines}}
    \label{fig:arp_4754b}
\end{figure}

The widely accepted paradigm for aerospace is to follow guidance from SAE ARP4761~\cite{arp47611996guidelines} and ARP4754A / ED-79A~\cite{arp4754a2010guidelines} (Figure~\ref{fig:arp_4754a}). This framework follows a classic systems engineering V-lifecycle (Figure~\ref{fig:arp_4754b}) to determine the level of certification required by system components. Using Functional Hazard Analysis (FHA), the system architecture is decomposed into different levels; from the aircraft level, down to subsystems. The functions of each level are then listed out and mapped to respective subsystems. The failure conditions and severity of each function then determine the Development Assurance Level (DAL) and establish the safety requirements to meet (Figure~\ref{fig:dal}). The applicant then follows a Means of Compliance (MOC) to verify that the failure condition requirements are met.

\begin{figure}[!ht]
    \centering
    \includesvg[inkscapelatex=false,width=\linewidth]{img/dal.svg}
    \caption{Assurance levels based on failure severity and probability~\cite{arp4754a2010guidelines}, p.14}
    \label{fig:dal}
    \vspace*{-2mm}
\end{figure}

DO-254 / ED-80~\cite{do2542000design} and  DO-178C / ED-12C~\cite{do178c2011software} are companion documents which provide guidance for the certification of hardware and software components, respectively. Similarly, DO-200B~\cite{do200b2015standards} provides guidelines for handling aeronautical data and could be extended to the management of machine learning data. For the scope of this paper, we focus on the software certification aspect. Following DO-178C, each requirement must be \textit{traceable} to multiple artifacts: the lines of code implementing it, the test cases verifying the correctness of that code, and the results of such tests. This ensures that each requirement is fulfilled by the source code, and that there is no extraneous code.

With machine learning models, mapping lines of code to requirements is not straightforward, hence creating a gap in the current certification process.
First, such models often contain millions of parameters, and subsets of those are activated depending on the properties of the input. This makes it impossible to trace the impact of a single parameter on the resulting output.
Second, those parameters are not directly derived from physics or requirements as is the case with traditional software (e.g., a Kalman filter, a yaw controller). The parameters are automatically learned from example data which is fed to the model during its training phase. Thus in order to trace this parameter to a requirement, that data itself should be mapped to the requirements. This shifts the focus of current certification processes from not only the code itself, but also towards data traceability.
Third, some machine learning models build intermediate representations of the data that are increasingly more complex, powerful, but also not interpretable. By looking at those intermediate layers in the code, it is not possible to map their parameters to requirements.

The previous observations prevent the bidirectional mapping that is required from requirements to code and parameters: \textit{additional means of compliance and standards are needed}. The next section describes attempts from key research and industry actors to fill this gap.

%% file: sections/01_related_work.tex
\subsection{Certification frameworks}
In their 2016 report on the verification of adaptive systems~\cite{faa2016verification}, the FAA first established some useful terminology, and drew the boundary between adaptive and non-adaptive systems. For the scope of this paper, we are concerned only with non-adaptive systems which do not change behavior in operation.

EASA then proposed an incremental Artificial Intelligence (AI) roadmap~\cite{easa2020roadmap, easa2023roadmap} focused on trustworthiness and the need for social acceptance. They propose ramping up capabilities from simpler crew assistance (2022-2025), to human/machine collaboration (2025-2030), and full autonomy (2035+). They identify a paradigm shift to focus the assurance process on the correctness and completeness of datasets, bias mitigation, model accuracy and performance, as well as novel verification methods.

\begin{figure}[!b]
    \centering
    \includesvg[inkscapelatex=false,width=\linewidth]{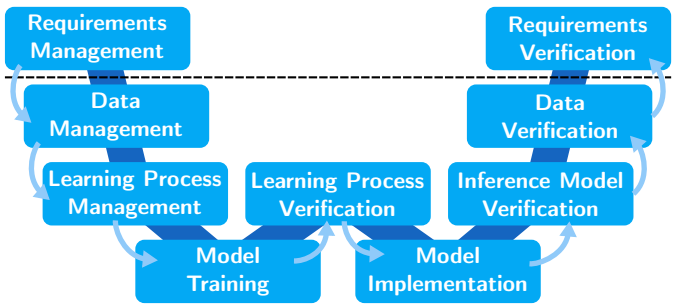}
    \caption{W-shaped cycle for learning assurance~\cite{easa2020codann}}
    \label{fig:w_diagram}
\end{figure}

EASA and Daedalean AG published the joint CoDANN report~\cite{easa2020codann} to study the \textit{learning assurance} concept introduced in the EASA roadmap. They present a novel W (double V) assurance cycle (Figure~\ref{fig:w_diagram}). In the first V, the model is trained and verified iteratively until a satisfactory performance is met. In the second V, the traditional verification process continues with the model implementation. For dataset validation, they present a \textit{distribution discriminator} to evaluate the completeness of datasets. The report establishes a distinction between robustness to training data (\textit{algorithm robustness}), and to input perturbations (\textit{model robustness}).

Shortly after, CoDANN II~\cite{easa2021codann2} addressed additional topics such as implementation, inference, and model explainability. CoDANN I and II are demonstrated on vision-based landing and Detect-and-Avoid (DAA) applications~\cite{faa2021autoland}, attesting to the modularity of the proposed framework. Following this work, EASA then published usable guidance for AI applications of level 1 \& 2 (i.e. human assistance), establishing clear objectives for applicants~\cite{easa2021concept, easa2023concept}.

\begin{figure}[!b]
    \centering
    \includesvg[inkscapelatex=false,width=\linewidth]{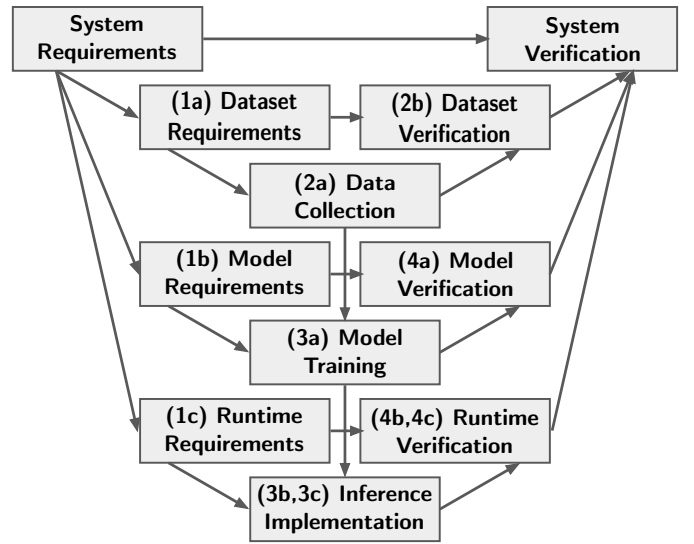}
    \caption{Modified V-diagram from~\cite{gariel2021framework}}
    \label{fig:v_diagram}
\end{figure}

Xwing Inc. proposed a modified V-diagram~\cite{gariel2021framework} in three stages: 1) data requirements and processes to verify data used to build and test the model, 2) model performance, and 3) reliability of operation at runtime (Figure~\ref{fig:v_diagram}).

Xwing presented results on a DAA application and compared a traditional radar-based approach with a computer vision method using machine learning. Their paper identifies the clear difference between the traditional physics-based bottom up approach, and the top-down approach required for machine learning applications. The authors emphasize that model parameters are just as important, if not more, than the code processing the network. They also propose to treat training data identically to aeronautical design data in traditional certification. However, the paper does not detail how to validate a machine learning model against performance requirements. We propose to extend this framework in our solution.

We can also note the contribution of working groups on the topic such as SAE G-34 and AFE 87. These groups are currently a work-in-progress and their initial findings echo the research aforementioned.

\subsection{Machine learning verification}

\subsubsection{Verification of Neural Networks (VNN)}
A major failure mode of ML models is local instability: a small input perturbation can result in large output variations. \textit{Adversarial attacks} exploit this vulnerability to trick models into giving incorrect outputs. VNN gives formal guarantees that a model is not susceptible to any attack within a given input range~\cite{liu2021algorithms}. While it is unlikely to adversely affect image formation at an airport or airspace level, robustness to small input variations should still be considered as they can occur in nominal operation.

Formal verifiers such as VENUS~\cite{botoeva2020efficient,kouvaros2021formal}, AI$^2$~\cite{gehr2018ai2} or VeriNet~\cite{henriksen2020verinet,henriksen2021bias} solve verification problems using a branch-and-bound approach. Counterexamples of unexpected behavior are obtained as a byproduct. However formal verifiers are slow and limited to small feed-forward architectures with ReLU activation functions, which fail to scale up to real-world aerospace models.

Reluplex~\cite{katz2017reluplex}, Marabou~\cite{katz2019marabou}, DeepTest~\cite{tian2018deeptest}, DeepZ~\cite{singh2018fast}, DeepSafe~\cite{gopinath2018deepsafe}, and DeepHunter~\cite{xie2019deephunter} all require knowledge of the model activations or architecture and are thus not scalable. Some of these approaches were successfully applied to the ACAS Xu aircraft collision avoidance algorithm~\cite{katz2017reluplex,katz2019marabou}. Most methods focus on pixel-space noise which can model realistic sensor errors, but fails to capture semantic errors  (e.g., brightness, saturation, scaling, etc.). Semantify-NN~\cite{mohapatra2020towards} and Tiler~\cite{yang2019correctness} provide robustness to families of semantic perturbations but modify the training process or require a full image formation model, which may prove impractical.\\

\subsubsection{Explainability}
A large body of work focuses on making neural networks interpretable by peeking into the hidden layers~\cite{linardatos2020explainable}. While core to the human-centric EASA roadmap, we consider it intractable due to the large and ever-changing number of model architectures, and possible variants. Model-agnostic interpretation techniques exist but still face many challenges~\cite{ribeiro2016should}. They do not provide the level of interpretability expected in traditional software certification but could still be useful for development purposes or post-mortem analyses: we will thus favor \textit{black-box} verification.\\

\subsubsection{Black-box verification}
\textit{Black-box} systems are opaque to the user, only the input and output can be observed, discarding the inner workings of the model. Corso et al.~\cite{corso2021survey} standardizes the problem formulation and studies applications to Autonomous Vehicles (AV) and aircraft collision avoidance. \textit{Adaptive stress testing} (AST)~\cite{lee2020adaptive} uses reinforcement learning to find likely system failure paths in simulation. The approach is limited to sequential decision systems and has not been generalized to AI models which process single images. By ranking failure scenarios, the simulation testing framework from~\cite{norden2019efficient} performed the first independent black-box evaluation of a commercial AV system.

The FAA considers the characteristics of training data as necessary conditions for verification~\cite{pham2022identifying}. Focusing on a black-box approach to neural networks, they propose the well-known $R^2$ indicator to determine the goodness-of-fit of a given model, and introduce the $S^2$ indicator to determine if the training set adequately covers the model operating range. While limited to a single neuron with no activation function, additional efforts are underway to extend the approach.

Failures can be rare and challenging to estimate in real flights (e.g., prohibitive number of tests~\cite{kalra2016driving} and dangerous scenarios), or in simulation (computationally expensive). NASA thus introduced SysAI~\cite{he2020framework} to efficiently generate simulated test cases using active learning and DynaTrees. This approach effectively uses less samples than Monte Carlo but requires parameterized boundary shapes for model failures, and is tailored to sequential data. Moss et al.~\cite{moss2023bayesian} then proposed a \textit{statistical verifier} which scales better than \textit{formal verifiers}. It uses Bayesian optimization and importance sampling to efficiently sample the Operational Domain (OD) as a parametric space. Using this minimal set of design points, a surrogate model is fitted for an inexpensive but robust estimate of the system failure probability. Results are successfully demonstrated on a simulated vision-based landing application.

%% file: sections/02_proposed_approach.tex
We choose not to open the AI black box. Indeed, certifying individual architectures would be ill-fated as their sheer number is intractable; architectures evolve rapidly in this very dynamic research field. Even the handful of well-established architectures are customized to optimize performance in use cases that differ for each applicant. Alternatively, a restriction to interpretable-only models would limit the capacity and performance of models that could be used, ultimately preventing the safety improvements offered by machine learning. As for \textit{explainability}, a deeper understanding of the inner layers of a model might be useful for development or debugging, but is unnecessary for the certification argument where the holistic performance of the system is evaluated. Similar to hardware qualification, this black-box approach is a key differentiator and shifts the paradigm from requirement-based code, to performance-based code.

Whether it takes the form of a W-diagram~\cite{easa2020codann} or a modified V-diagram~\cite{gariel2021framework}, the research community has converged on three-stage verification: dataset, model, and inference. We follow this decomposition and start by establishing performance requirements for the ML subsystem.

\subsection{Requirements development}  \label{sec:requirements}
We follow the traditional ARP4761 and ARP4754 standards in place. The applicant should first define a \textit{complete} and \textit{correct} Operational Design Domain (ODD) of the system: an exhaustive list of conditions in which the model will operate, and their expected proportions. For autonomous aircraft, examples can be flight location,  weather conditions, time of day, runway type, etc.

\begin{figure}[!ht]
    \centering
    \includesvg[inkscapelatex=false,width=\linewidth]{img/fha.svg}
    \caption{Notional functional hazard analysis with ML function highlighted (yellow)}
    \label{fig:fha}
    \vspace*{-2mm}
\end{figure}

The traditional FHA and Fault Tree Analysis (FTA) are then carried out. As a result, the ML function is assigned a DAL level, and an associated probability of failure $P_\text{fail}$ (Figure~\ref{fig:fha}). 
EASA recognizes that estimating $P_\text{fail}$ is necessary but not sufficient on its own, and additional model performance artifacts are needed for certification (see~\cite{easa2023concept}). For the scope of this paper, we present a methodology to substantiate the $P_\text{fail}$ computation. The functional decomposition above helps isolating the behavior of the ML component and sets its context. As can be seen in Figure \ref{fig:vbl_architecture}, it is a small function of a more comprehensive system and does not operate the controls of the aircraft directly. The output of the model is bounded and monitored by safety layers and other non-ML subsystems which follow the traditional certification process.

The problem is then reduced to two things: 1) following software development processes consistent with the DAL level, and 2) computing the probability of failure for the ML system over the ODD (Figure~\ref{fig:steps}).

\subsection{Dataset validation and verification}
The data is split between independent training, validation, and test sets which identically represent the ODD (i.e. drawn from the same distribution). The training and validation sets are used by developers to optimize respectively weights and hyperparameters of the model. The test set or \textit{certification set} is kept separate by a quality or certification team. Never seen by developers, this set ensures that good model performance actually stems from unbiased \textit{generalization} on unseen data, rather than overfitting the model to the training/validation data.

\begin{figure}[!t]
    \centering
    \includesvg[inkscapelatex=false,width=\linewidth]{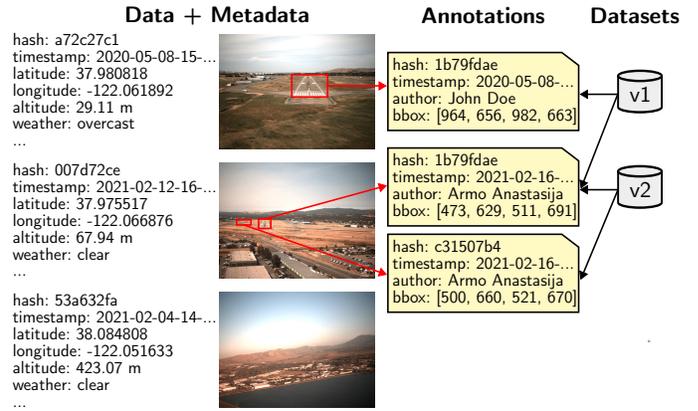}
    \caption{Data version control. Data and metadata (here images and labels) are hashed into a repository to keep track of their integrity. Each datum can be assigned multiple labels (or no label). Versioned data and metadata can then be used to assemble datasets for training or testing algorithms.}
    \label{fig:dvc}
\end{figure}

These datasets evolve continuously: applicants might add additional data focusing on model failures, correct ground truth labels, or remove outdated data (e.g., new sensor). For complete traceability, the applicant should track these changes as Engineering Change Orders (ECO) in a Product Lifecycle Management (PLM) system. This version control applies to the data itself, but also to the associated metadata: ground truth labels, timestamp, geolocalization, aircraft attitude, and other relevant descriptions (Figure~\ref{fig:dvc}). Those attributes can then be used to analyze the completeness of the operational domain. The traceability of the certification set is especially important: the applicant should validate that no examples from that set were used in developing the model. By versioning the changes, it is also always possible to reproduce previous models. We recommend the same traceability for open-source pre-trained models, usually developed by large corporations on very expensive cloud infrastructure. In theory, this would enable reproducing a given model. Synthetic data should follow the same process and any simulator should be validated through flight test in order to minimize any simulation-to-reality gap (i.e., the \textit{sim2real} problem).

Finally, dataset verification also ensures that acceptable performance on the test set translates to acceptable performance in the operational domain. As such, the test set should capture the variations observed in the ODD, and span the parameter ranges that the applicant defined in their parametric decomposition. Rare corner cases should be also included to test the robustness of the model and weighted according to their operational likelihood. To verify the coverage of the test dataset, we propose to use the $R^2$ and $S^2$ indicators~\cite{pham2022identifying}, and the distribution discriminator~\cite{easa2020codann}.

\subsection{Training verification}
We focus on parametric supervised learning models which are deterministic: for a given input, the model will always yield the same output. We do not consider \textit{online learning} because the model would keep learning after being deployed into operation, and inference would run on an unverified model.

Model iterations are often required to keep on improving a baseline performance. Each experiment might use different data, hyperparameters, model architecture, or training process. When a model is trained, a number of artifacts need to be tracked in order to ensure experiments are reproducible (Figure~\ref{fig:steps}):
\begin{itemize}
    \item The data and labels used to train the model (including training/validation split)
    \item Model architecture, hyperparameters, and weights
    \item Training hardware and software (including versioned dependencies)
    \item Model initialization weights (before training)
    \item Seeds used for pseudo-random number generators
    \item Training logs and metrics
\end{itemize}

For certification, the applicant should implement traceability of at least: the artifacts necessary to instantiate the model (architecture, hyperparameters, weights), and the certification dataset. Akin to hardware certification, we argue that these two artifacts alone are sufficient to build the model and verify that it passes the robustness and performance tests on the ODD. Many different models could be designed from the same training/validation data. Following our black-box approach, we argue that traceability of other artifacts and training/validation datasets is a healthy and encouraged internal practice when developing the models, but might not be as critical for certification substantiation.

\subsection{Inference verification}
Once a suitable model is obtained in the development phase, it has to be ported as executable code on the target hardware. This porting exercise is usually necessary because aircraft computers differ in requirements (e.g., real-time constraints, memory, etc.) from the development computers used to train the models. This model conversion can slightly modify the original research model (layers grouping, weight quantization, pruning, hardware threading determinism, etc.) and impact detection performance. Inference software and hardware can be developed with traditional means of compliance such as DO-178C and DO-254. The ported model can now be considered as a black-box system to verify. Inference is verified in terms of performance and operational runtime guarantees.

For performance, we need to compute the probability of failure $P_\text{fail}$ (see section \ref{sec:requirements}). To this end, we center our framework around the \textit{Bayesian safety validation (BSV)} algorithm~\cite{moss2023bayesian} to efficiently sample the ODD and obtain an accurate estimate of $P_\text{fail}$. The applicant should define metrics that evaluate the performance of the model over its expected tasks and operational domain. For machine learning models, some common metrics include: mean average precision (mAP), precision and recall, false positive rate (FPR), mean squared error (MSE), mean absolute error (MAE), etc. These metrics depend on the type of task (e.g., classification, regression, tracking) and should be defined by the applicant.

As for runtime requirements, they can be verified and validated using traditional methods. For machine learning models, this step ensures that the model obtained in the research phase is correctly ported to the target hardware. Again, the traceability of the model architecture, weights, and hyperparameters is essential to verify this step. Other aspects to be verified include: throughput, latency, execution errors, determinism. It is important to not that these requirements do not apply to the development/training code.

%% file: sections/03_results.tex
This section introduces the use case of vision-based landing, before showing how we apply the proposed approach.

\subsection{Vision-based landing}
\begin{figure}[!t]
    \centering
    \includegraphics[width=\linewidth]{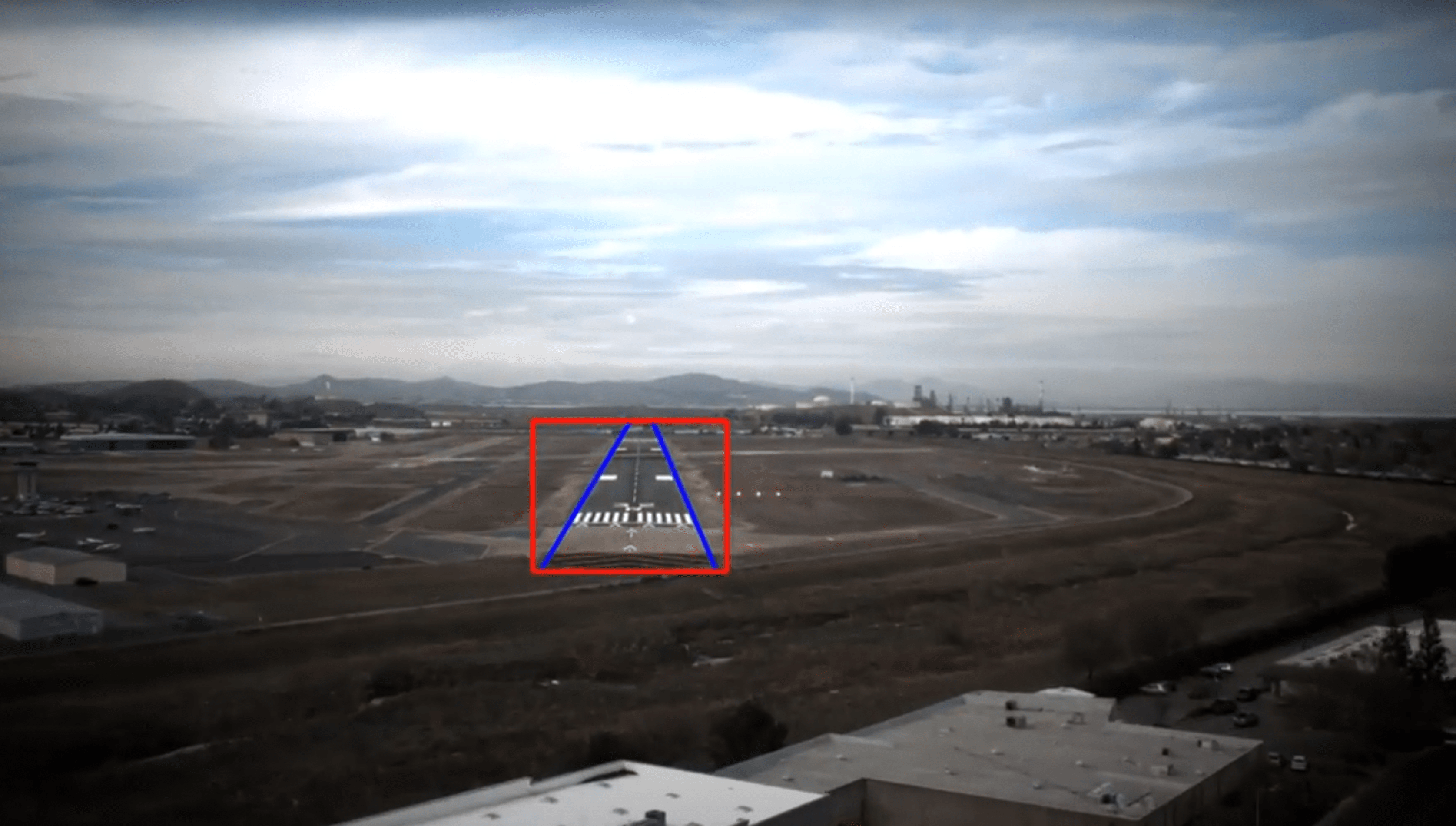}
    \caption{Vision-based landing using a machine learning component to detect a runway. Runway corners and sidelines are detected in blue, defining a coarse bounding box in red.}
    \label{fig:vbl_image}
    \vspace*{2mm}
\end{figure}

\begin{figure}[!b]
    \vspace*{6mm}
    \centering
    \includesvg[inkscapelatex=false,width=\linewidth]{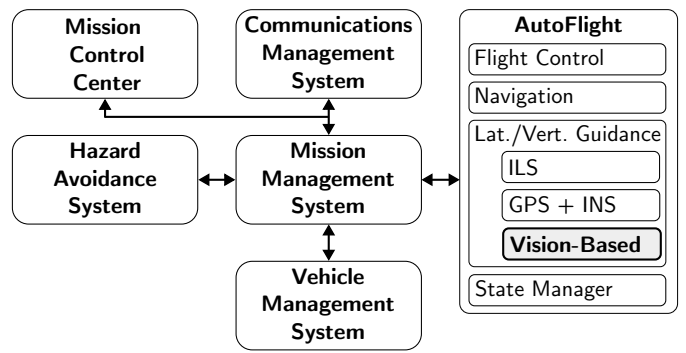}
    \caption{Notional autonomous flight components. The vision-based subsystem is a small part of the comprehensive system.}
    \label{fig:mms_architecture}
\end{figure}

A common application in autonomous flight, Vision-Based Landing (VBL) (Figure~\ref{fig:vbl_image}) uses only camera images to augment the landing guidance obtained from GPS and/or Instrument Landing System (ILS). This is a typical example of computer vision application that was enabled by the resurgence of deep learning. Looking at the high-level architecture of a notional autonomous flight system (Figure~\ref{fig:mms_architecture}), we find interconnected components from the high-level mission management system, down to the flight controls. The low-level flight control components help make the system \textit{independently safe}~\cite{kimchi2019independently}: they maintain the flying state of the aircraft and its safety, before taking into account other inputs from high-level components.

\begin{figure}[!ht]
    \centering
    \includesvg[inkscapelatex=false,width=\linewidth]{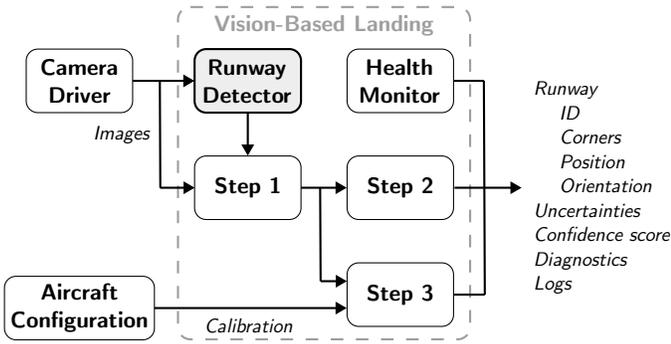}
    \caption{Notional VBL architecture. The machine learning model is used only for one step of the overall algorithm.}
    \label{fig:vbl_architecture}
\end{figure}

One of the systems is in charge of vertical and lateral guidance to the runway. In that subsystem, vision-based landing provides one solution, along with ILS and GPS solutions. In the VBL block, an ML component is used to detect the runway in the image, while other traditional computer vision algorithms perform tracking across frames, and pose estimation (Figure~\ref{fig:vbl_architecture}). This separation of concerns helps understand where the ML model fits in the overall system, and visualize how it constitutes only a small portion of the complete solution.

The input to the VBL system is an image with the associated sensor calibration (intrinsics and extrinsics), the main output is the position and orientation of the aircraft with respect to the runways detected in the image. The different components can be described as follows:
\begin{itemize}
    \item \textbf{Camera driver:} Acquires images from the sensor and feeds them to the rest of the pipeline.
    \item \textbf{Runway detector:} The ML component. A model trained to detect runways in images. For each runway detected in the image, the model outputs the four runway corners position in the image, as well as uncertainties on the prediction. Many convolutional neural network (CNN) architectures can implement this task: single-shot detectors, region-proposal networks, anchorless networks, or more recently transformers. Note that machine learning algorithms other than CNNs can also be used such as Histogram of Gradients (HoG) combined with Support Vector Machines (SVM) or Random Forests.
    \item \textbf{Pose estimator:} Uses the corners of the runway and camera sensor calibration to compute the position and orientation (i.e., \textit{pose}) of the camera (and aircraft) with respect to the runway.
    \item \textbf{Tracker:} Uses traditional a computer vision approach (i.e., non ML-based) to track the corners of the runway across frames and provide robust position estimates.
    \item \textbf{Health monitor:} Runs consistency checks on the output of the previous steps: Is the prediction confidence high enough? Is this the expected runway? Are the runway corners geometrically consistent? Is there a large jump in position estimate?
\end{itemize}

Note that the output of the machine learning model is gated and validated in many ways: by its own confidence score, by the geometric consistency of its output, by the VBL health monitor, by comparison with other estimators (ILS, GPS+INS), by the mission management system, and finally the vehicle management system. Additionally, each system has redundancies and consistency checks that can be run on the output of several VBL algorithms running in parallel.

\subsection{Verification}
\subsubsection{Requirements development}
We carry out the FHA and FTA of the system and its components (Figure~\ref{fig:vbl_architecture}), and arrive at a DAL level C with an associated failure probability of $10^{-4}$ per approach over the operational domain (Figure~\ref{fig:fha}), for the runway detector model.

\begin{figure}[!t]
    \centering
    \includesvg[inkscapelatex=false,width=\linewidth]{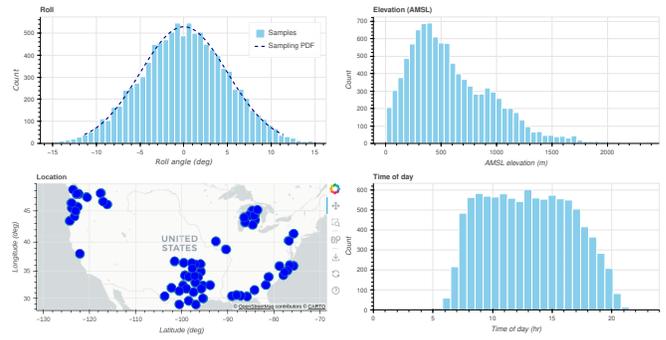}
    \caption{Some example ODD dimensions after a draw of 10,000 samples (roll angle, elevation, location, time of day)}
    \label{fig:distributions}
\end{figure}

Given that the runway detector is only active during the approach phase of the flight, we condition our probability distribution on the approach: $P(\text{fail} \mid \text{approach})$. We then parameterize our operational domain along known dimensions of variability. For pedagogic and plotting purposes, we decompose our ODD along only two continuous dimensions: glide slope angle $\alpha \sim \mathcal{N}\left(3, 0.3, \left[1, 7\right]\right)$ degrees, and distance to the runway $d \sim \mathcal{N}\left(0,1.5,\left[0,4\right]\right)$ nautical miles. Note that $\mathcal{N}\left(\mu,\sigma,\left[a,b\right]\right)$ represents the truncated normal distribution with mean $\mu$, standard deviation $\sigma$, and truncation interval $\left[a,b\right]$. Additional dimensions are presented on Figure~\ref{fig:distributions} for example.

\subsubsection{Dataset verification}
We use data version control to track changes to our data and metadata. Datasets are assembled from versioned data and metadata and used to train and test the models. The integrity of the data is guaranteed by computing a hash, a unique digital signature that depends on the content of the data itself. If the data is edited, or corrupted in any way, the hash also changes.

Runway images are labeled in batches through third-party providers. Annotators are given precise instructions on how to label a runway and each batch is subject to quality analysis on a random sample. A runway label consists of annotating the full extent of visible runway edges, from which we can derive runway corners for example (Figure~\ref{fig:vbl_image}).

\begin{figure}[!ht]
    \centering
    \includegraphics[width=\linewidth]{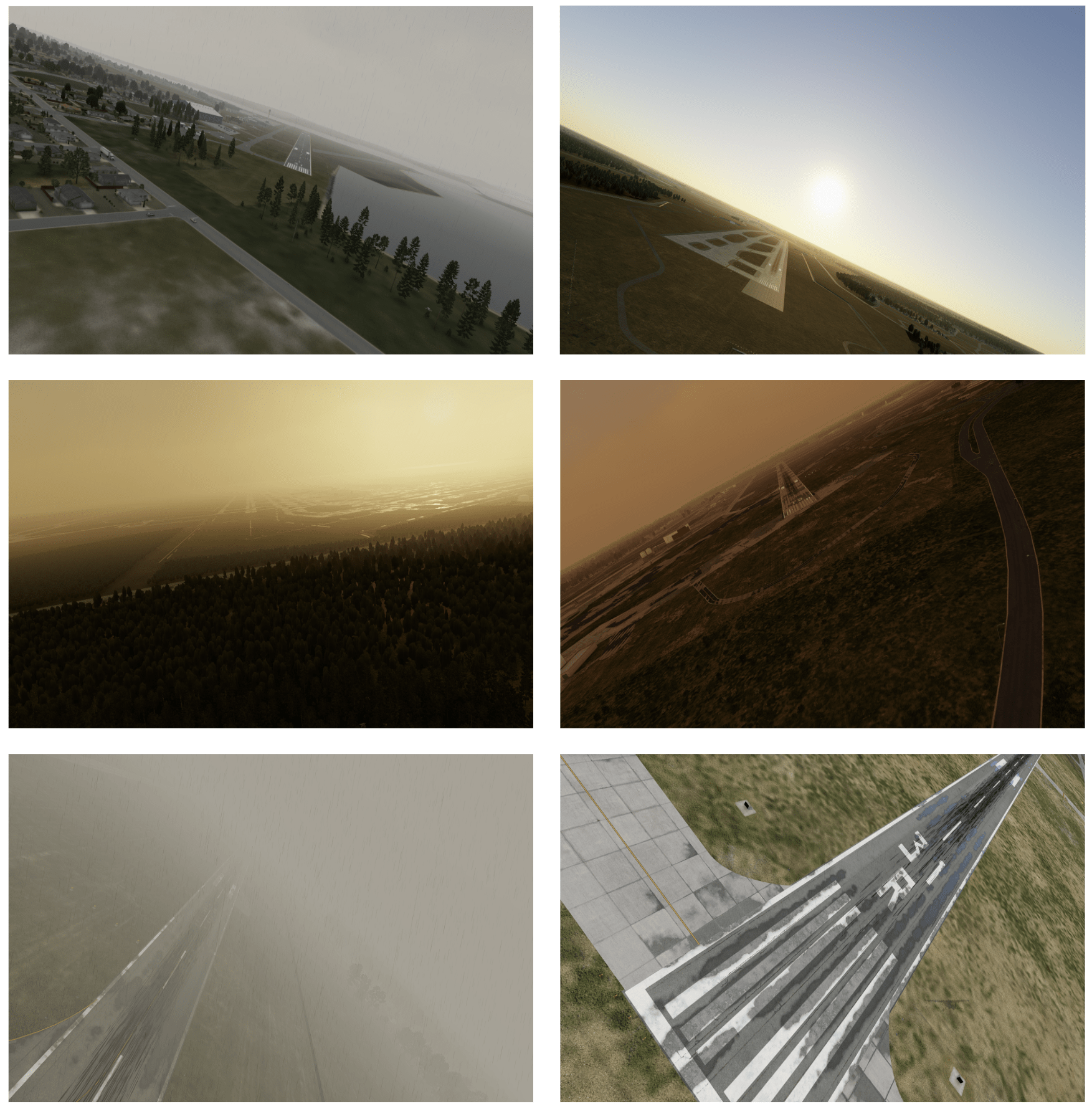}
    \caption{Synthetic images covering different conditions of the parametric space (sun in the field of view, weather, dangerous aircraft position and orientation, etc.)}
    \label{fig:synthetic}
\end{figure}

We augment our dataset by using synthetic images generated by a physics-based rendering engine (Figure \ref{fig:synthetic}).

For independence of the different sets, we use a special splitting policy so that no two images from the same landing sequence end up in both the training and the certification set. In addition to version control, we also propose a strict data access policy. We propose that every data collection campaign be gated by an internal certification team or in collaboration with Designated Engineering Representatives (DER). A given amount of data is kept to contribute to the certification test set, while the rest is released to the engineering team to develop the models. Keeping this separation ensures that no practitioner can cheat and boost model performance by accessing the certification test data while developing the model. As part of the development cycle, the model is then provided to the certification team or DERs to be run on the certification data, and the evaluation results are returned to the engineering team for approval or further developments.

We propose \textit{healthy} operational overfitting by training on all the runways where the aircraft is expected to be deployed. This includes 65 airports where Xwing Air\footnote{Xwing Air is the cargo airline of Xwing, Inc. Under a Part 135 certificate, their fleet of 35 Cessna Caravans operates more than 400 weekly flights to shuttle packages across 34 UPS delivery routes.} operates (Figure~\ref{fig:distributions}). This set is extended to also include alternate airports in case of diversion. This \textit{healty overfitting} ensures that any runway encountered in operation has been seen by the model in training, and should result in good detection performance. This set represents 80\% of the total development data and is completed by additional airports and runways to promote performance generalization further. Note that the test set or certification set consist only of real-world data from the operational domain and is not accessed by developers during the training phase. Using a common 90/10\% training/test split, a breakdown is presented in Table~\ref{tab:dataset}.

\subsubsection{Model verification}
We use the PyTorch deep learning framework in deterministic mode to develop our models. For every training session, we take a snapshot of the code, its software dependencies, and the hardware configuration. This ensures that the training is fully deterministic and repeatable.

\begin{table}[!ht]
\centering
\begin{tabular}{lrr}
\textbf{Split} & \textbf{Flights} & \textbf{Images}\\
\hline
\textbf{Development} & & \\
~Training (real) & $45$ & $23{,}442$ ($65\%$)\\
~Training (synthetic) & N/A & $5{,}000$ ($14\%$)\\
~Validation (real) & $10$ & $4{,}067$ ($11\%$)\\
\hline
\textbf{Certification} & & \\
~Test (real) & $9$ & $3{,}488$ ($10\%$)\\
\hline
\textbf{Total} & $65$ & $35{,}997$ ($100\%$)\\
\end{tabular}
\caption{\label{tab:dataset}Dataset composition}
\end{table}

We use \textit{transfer learning} (i.e. pre-trained weights) to leverage proven state-of-the-art models for single-class (i.e. runway) object detection. Our model stems from a Keypoint R-CNN architecture with a ResNet-50-FPN backbone~\cite{he2017mask}. The ResNet backbone acts as a task-agnostic feature extractor and is pre-trained on the very large corpus of the ImageNet dataset~\cite{deng2009imagenet}. The keypoint detector is pre-trained on the COCO 2017 Keypoint dataset~\cite{lin2014coco}. The model outputs a bounding box around the runway, a confidence score, as well as keypoints of interest (the corners of the runway). The loss minimized during training is a combination of bounding box regression loss (position, size), and Object Keypoint Similarity (OKS).

\begin{figure}[!b]
    \centering
    \includesvg[inkscapelatex=false,width=\linewidth]{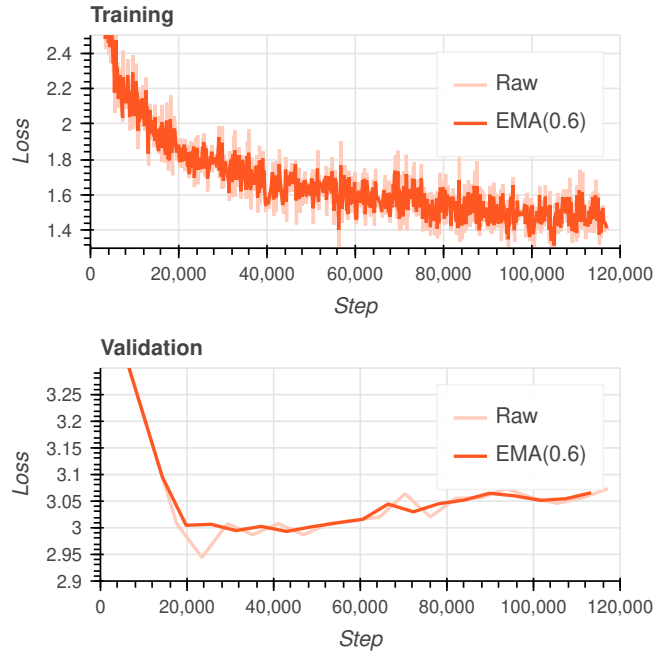}
    \caption{Learning curves plotting the loss function per optimizer iteration (mini-batch). The checkpoint around iteration $50{,}000$ achieves best generalization performance on the validation set.}
    \label{fig:learning_curves}
    \vspace*{4mm}
\end{figure}

We minimize the loss function with a Stochastic Gradient Descent (SGD) optimizer with a momentum of $0.9$ and a maximum of $90{,}000$ iterations (Figure~\ref{fig:learning_curves}). We feed $4$ images per mini-batch with a base learning rate $lr = 0.02$. For warmup and scheduling, we decay $lr$ by $\gamma = 0.1$ after $30{,}000$, $60{,}000$, and $80{,}000$ iterations. The model is trained with on-premise deep learning infrastructure. During training, we apply random image augmentations (e.g., blur, color jitter, rotation, scale, fog, rain, snow, sunflare, etc.) to prevent overfitting and improve the robustness of the network. At the end of model development, we choose the model around iteration $50{,}000$ which minimizes the validation loss; using \textit{early stopping} as a form of regularization to minimize overfitting (Figure~\ref{fig:learning_curves}).

\subsubsection{Inference verification}
Confidence scores below $0.7$ are discarded when computing predictions. We track two main metrics for the runtime performance of the model:
\begin{itemize}
    \item $AP_{BB}$: The average bounding box precision. Computed for Intersection Over Union $IoU \in \left[0.50,0.95\right]$ for all object sizes. We limit the number of model detections to a maximum of $100$.
    \item $AP_{KP}$: The average keypoints precision. Computed for Object Keypoint Similarity~\cite{lin2014coco} $OKS \in \left[0.50,0.95\right]$ for all object sizes. We limit the number of model detections to a maximum of $20$.
\end{itemize}

After a few development iterations and hyperparameter adjustments, the best and final model achieves $AP_{BB} = 0.89$ and $AP_{KP} = 0.99$ on our validation set. We convert the model from the developmental PyTorch format to an inference executable using runtime libraries such as NVIDIA TensorRT, Apache TVM, or ONNX. The inference engine compiles the model into an optimized set of instructions compatible with GPUs. By tracing model architecture, weights, and hyperparameters (Figure~\ref{fig:steps}), we can easily verify that the executable corresponds to the ported model. Note that the porting exercise might perform some weight quantization. This is acceptable since the conversion is deterministic, and the final certification metrics and performance are computed on the final inference model.

\subsubsection{Requirements verification}
\begin{figure}[!t]
	\centering
        \includesvg[inkscapelatex=false,width=\linewidth]{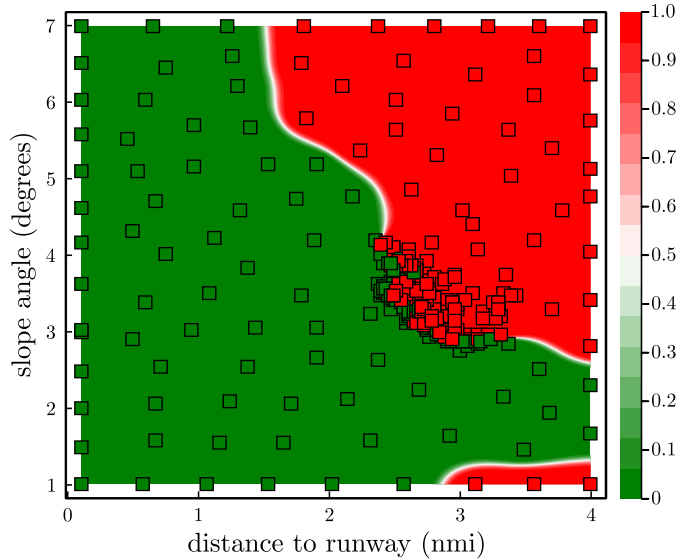}
	\caption{Probabilistic failure surrogate from an example iteration of the BSV algorithm on the runway detector. The algorithm efficiently samples (the squares) the ODD (distance, glideslope) to find the failure boundary (white). The runway detector failures are represented in red, successes in green.}
	\label{fig:pfail_vbl}
\end{figure}

Using our inference model, we run the probabilistic safety validation algorithm and obtain a failure boundary similar to that of Figure~\ref{fig:pfail_vbl}. For the sake of the example, we stop the BSV algorithm early at only 33 iterations, yielding an estimated failure probability of the ML subsystem is $P_\text{fail} = 5.8 \times 10^{-3}$. By placing ourselves in a situation where performance does not meet the initial $10^{-4}$ requirement, we can now illustrate example follow-up steps for applicants:

\begin{itemize}
    \item Refining the ODD: we can decide to revise the subsystem requirements or concept of operations (CONOPS) to restrict the model to regions where it performs well. In our example, the applicant could decide to only activate the runway detector below 2 nautical miles from the runway to now meet the $10^{-4}$ $P_\text{fail}$ requirement.
    \item Improving the model: BSV highlights failure regions which can be used to update the model to perform better in these areas. To prevent simply overfitting the certification dataset, the data access policy should prevent the counter-examples themselves from being used directly to re-train the model. Rather, general characteristics of those failure modes and regions could be communicated to developers.
    \item Validating the surrogate: the BSV generates a minimal set of test points which can be collected with real-world flight test data to validate the simulator, and the BSV failure boundary.
\end{itemize}

This example highlights the iterative nature of the machine learning model development. We repeat all the steps of the process, keeping track of the different experiments and artifacts, until a satisfactory performance is met.

%% file: sections/04_conclusion.tex
Safety is an emergent system property, not a component property: we considered a machine learning model as a small component of a more comprehensive autonomous flight stack. This hierarchical decomposition bounds the output of the model with  many redundancies and safety layers. The stack follows the conventional certification process for traditional software subsystems. For the machine learning subsystem, we propose a model-agnostic and tool-independent framework applicable to a wide variety of use cases, data types, and machine learning architectures.

Throughout a modified V-diagram, we emphasize the traceability of data, metadata, and model artifacts. After defining a complete operational domain, we sample simulated test cases to run through the model. We present the first use of the \textit{Bayesian safety validation (BSV)} algorithm~\cite{moss2023bayesian} in a complete certification argument. This statistical verifier scales to the large architectures currently used in research and industry. As opposed to other verification approaches, we sample the operational domain (e.g., distance to runway, weather), not the sensor space (e.g., pixel color, brightness). The verifier converges on an accurate estimate of the probability of failure $P_\text{fail}$ of the ML component, which we can tie back to requirements, closing the ML certification gap. This work is currently being used to supplement the FAA certification process of an autonomous cargo aircraft.